\documentclass[lnicst]{svmultln}
\usepackage{makeidx}
\usepackage{epsf}
\usepackage[dvips]{color}
\begin{document}

\mainmatter

\title{Approaching the linguistic complexity}

\titlerunning{Linguistic complexity}

\author{Stanis{\l}aw~Dro\.zd\.z\inst{1,2} \and
Jaros{\l}aw~Kwapie\'n\inst{1} \and Adam Orczyk\inst{1}}

\authorrunning{Stanis{\l}aw~Dro\.zd\.z et al.}

\institute{Polish Academy of Science, Institute of Nuclear Physics,
31-342 Krak\'ow, Poland, \\
\email{Stanislaw.Drozdz@ifj.edu.pl}
\and
University of Rzesz\'ow, Faculty of Mathematics and Natural Sciences,
35-310 Rzesz\'ow, Poland}

\maketitle

\begin{abstract}

We analyze the rank-frequency distributions of words in selected English
and Polish texts. We compare scaling properties of these distributions in
both languages. We also study a few small corpora of Polish literary
texts and find that for a corpus consisting of texts written by different
authors the basic scaling regime is broken more strongly than in the case
of comparable corpus consisting of texts written by the same author.
Similarly, for a corpus consisting of texts translated into Polish from
other languages the scaling regime is broken more strongly than for a
comparable corpus of native Polish texts. Moreover, based on the British
National Corpus, we consider the rank-frequency distributions of the
grammatically basic forms of words (lemmas) tagged with their proper part
of speech. We find that these distributions do not scale if each part of
speech is analyzed separately. The only part of speech that independently
develops a trace of scaling is verbs.


\end{abstract}

\section{Introduction}

Even though central to the contemporary science the concept of complexity
- by its very nature - still leaves its precise definition a somewhat open
issue. In intuitive and qualitative terms this concept refers to diversity
of forms, to emergence of coherent and orderly patterns out of randomness,
but also to a significant flexibility that allows switching among such
patterns on a way towards searching for the ones that are optimal in
relation to environment. Physics offers the methodology and concepts that
seem promising for formalizing complexity. One of those concepts is
criticality implying a lack of characteristic scale which indeed finds
evidence in abundance of power laws and fractals in Nature.

Whatever definition of complexity one however adopts, the human language
deserves a special status in the related investigations. It not only led
humans to develop civilization but it also constitutes - from a scientific
perspective - an extremely fascinating and complex dynamical
structure~\cite{nowak00}. Like many natural systems, language during its
evolution developed remarkable complex patterns of behaviour such as
hierarchical structure, syntactic organization, long-range correlations,
and - what is particularly relevant here - a lack of characteristic scale.
This latter phenomenon - in quantitative linguistics commonly referred to
as the Zipf law - describes the rank-frequency distribution of words in a
(sufficiently large) piece of text. This well-known, quantitatively
formulated in 1949 by G.K. Zipf observation~\cite{zipf49}, based
originally on ``Ulysses'' and latter on confirmed for many other literary
texts, states that frequency of the rank-ordered words is inversely
proportional to the words' rank. It needs to be added here that this law
constitutes a principal reference in quantitative linguistics and
inspiration for ideas and development in many different areas of science.

Zipf suggested interpretation of this law in terms of the so-called
principle of least-effort~\cite{zipf49,ferrer03}. This interpretation was
however soon questioned after it had been shown that the Zipfian relation
applies also to a ``typewriting monkey'' example~\cite{miller57}, i.e. an
essentially purely random process. This pointed to the Zipf law as too
indiscriminate to reflect the complex organization of languages.

\section{Results and discussion}

\begin{figure}[t]
\hspace{2.0cm}
\epsfxsize 8cm
\epsffile{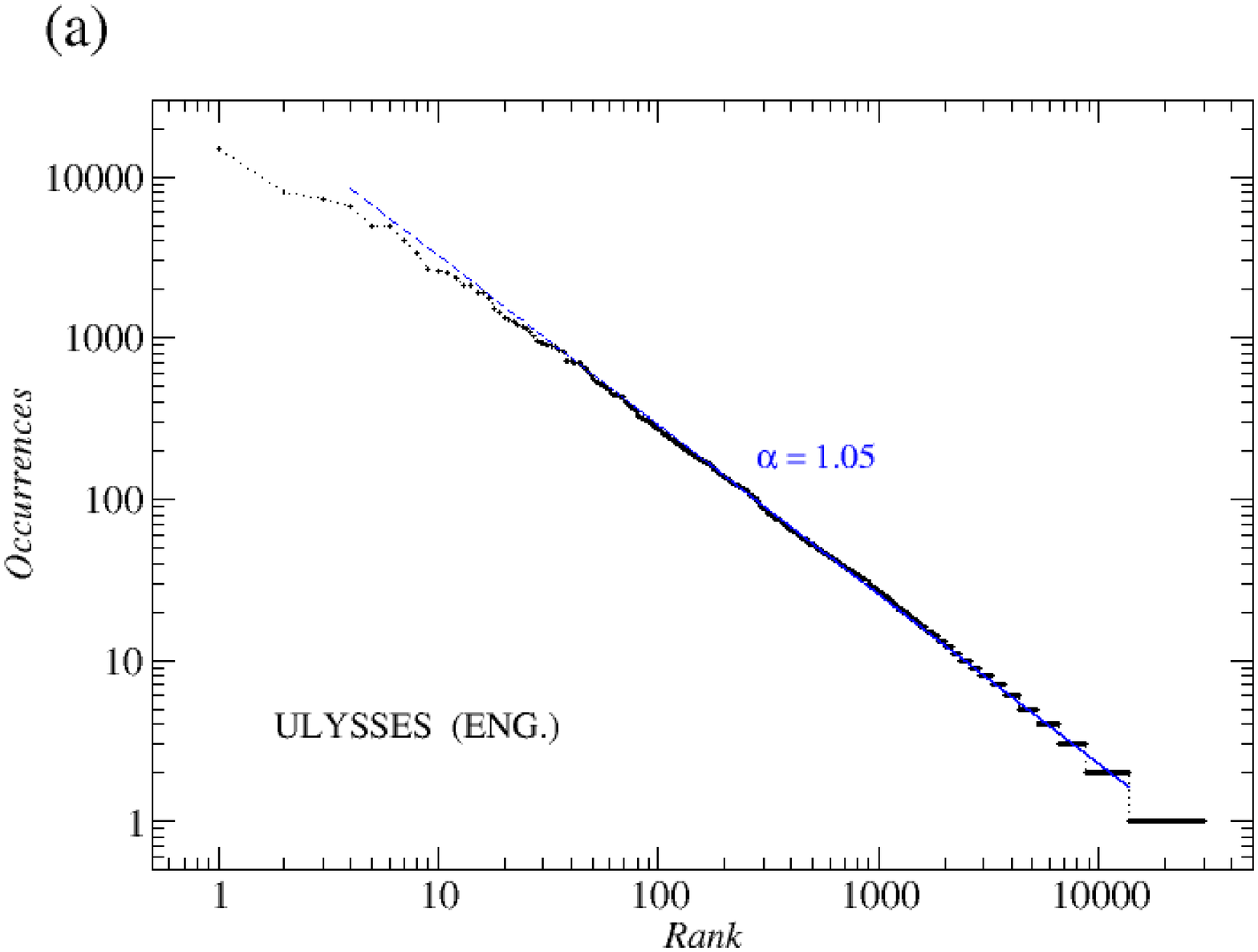}

\vspace{-0.5cm}
\hspace{2.0cm}
\epsfxsize 8cm
\epsffile{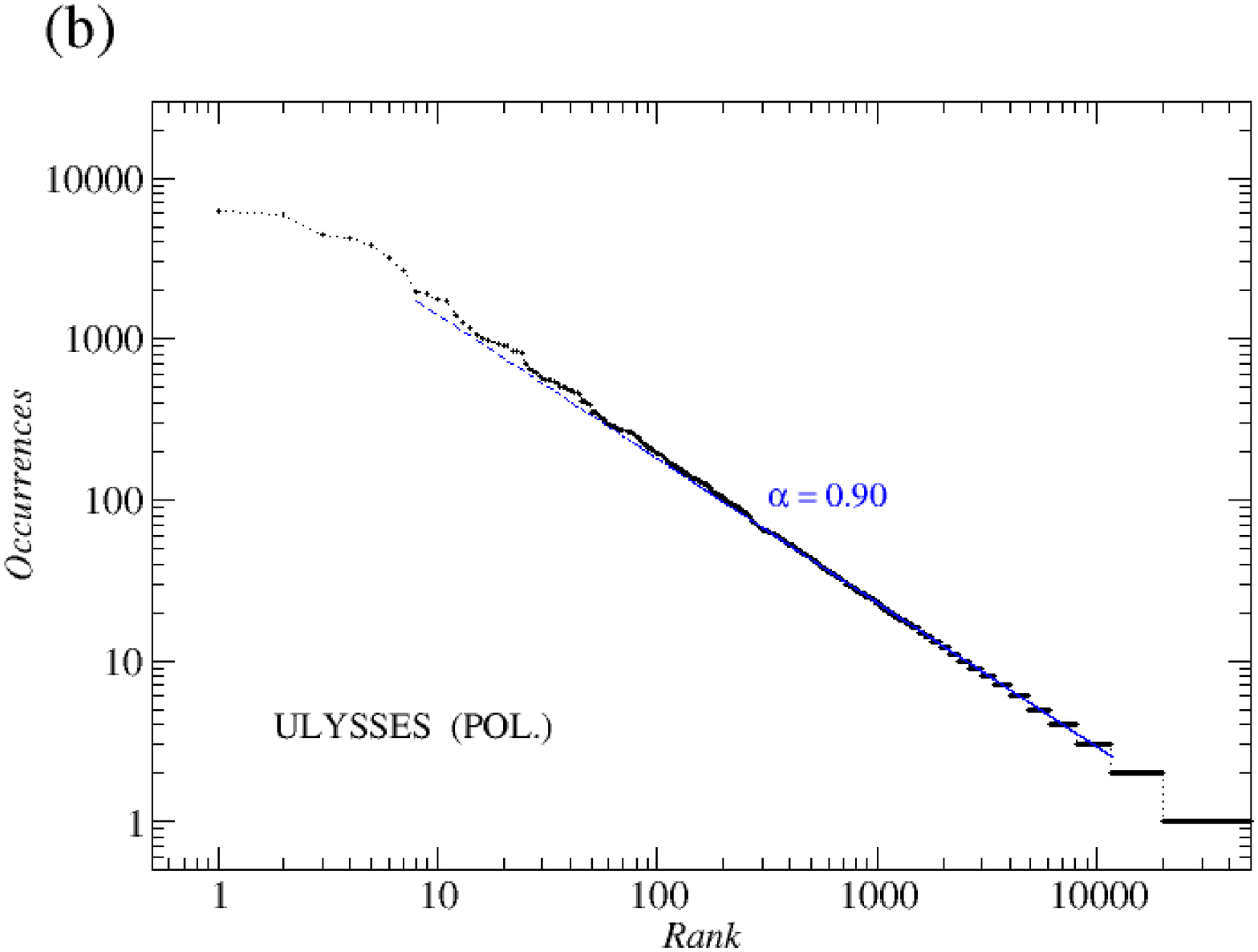}

\vspace{-0.5cm}
\hspace{2.0cm}
\epsfxsize 8cm
\epsffile{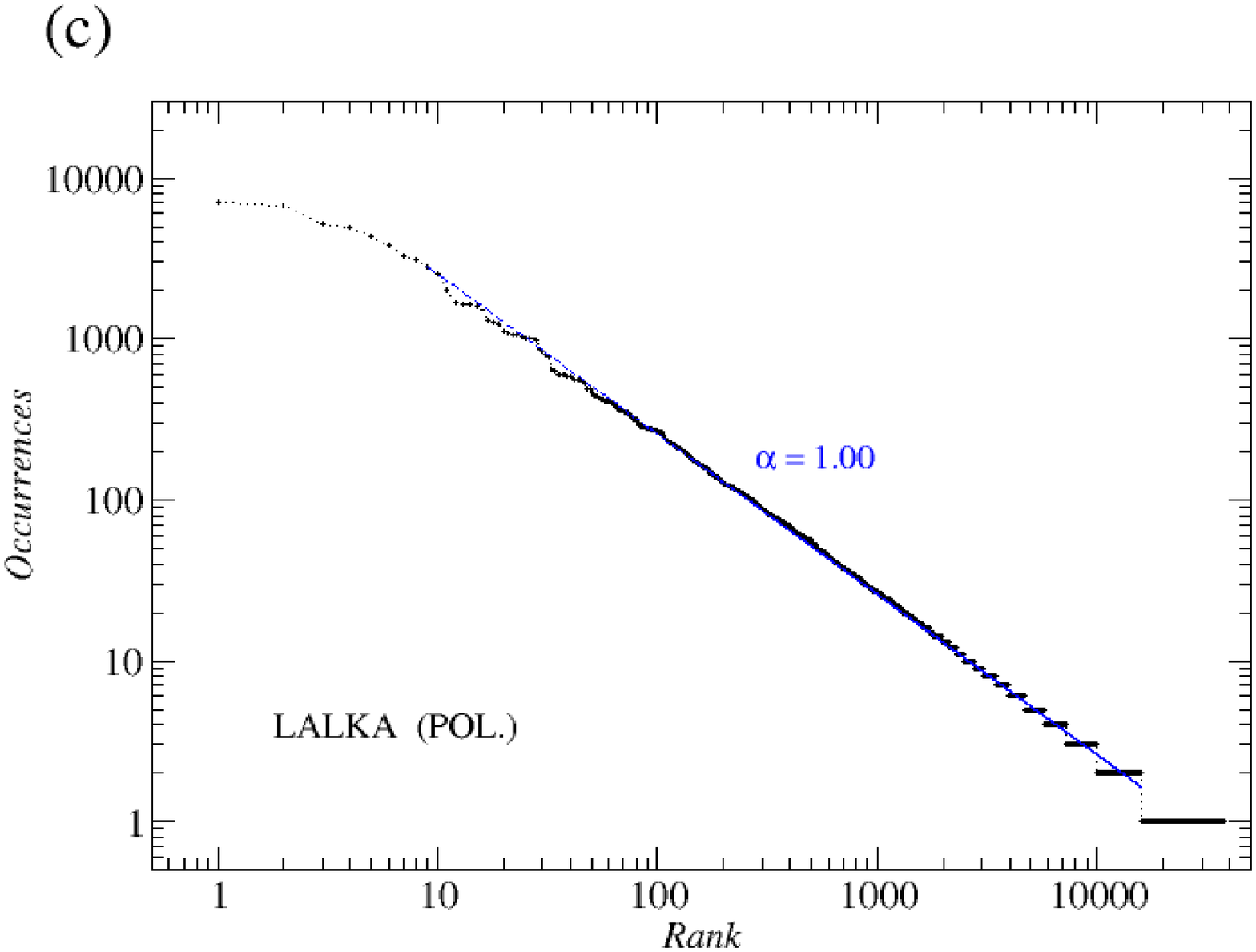}
\caption{Rank-frequency distributions of words in the English (a)
and Polish (b) text of ``Ulysses'' by James Joyce, as well as in the
native Polish text of ``Lalka'' by Boles{\l}aw Prus. All the three texts
have comparable lengths of 240,000-260,000 words.}
\end{figure}

Our recent studies based on English as well as on Polish texts open a new 
perspective to comprehend the linguistic complexity and sheds a new light 
on an involved message encoded in the Zipf law. First, referring to the 
classic analysis of ``Ulysses'' by James Joyce, carried out by 
G.K.~Zipf~\cite{zipf49}, we compare properties of the rank-frequency 
distributions of words in the English and the Polish 
text~\cite{slomczynski} of this novel. Results are shown in Figure 1(a) 
and Figure 1(b). Both versions of the text show scaling behaviour over 
three decades between the ranks 10 and 10,000. However, the scaling 
exponent for the Polish text ($\alpha \simeq 0.90$) is significantly 
smaller than for the English original ($\alpha \simeq 1.05$). This 
difference might originate from a far more inflectable character of the 
Polish language, which demands a larger set of words (understood as 
particular sequences of letters) to reproduce the course of narration in 
``Ulysses''. This observation may be considered a regularity, since 
typical Polish texts have smaller $\alpha$ than typical English texts. 
There are, however, Polish texts which have a value of $\alpha$ that is 
similar to English standards as it is documented in Figure 1(c). It is 
noteworthy that each of the three texts are well approximated by power-law 
distributions almost over the whole range of ranks.

\begin{figure}[t]
\hspace{2.0cm}
\epsfxsize 8cm
\epsffile{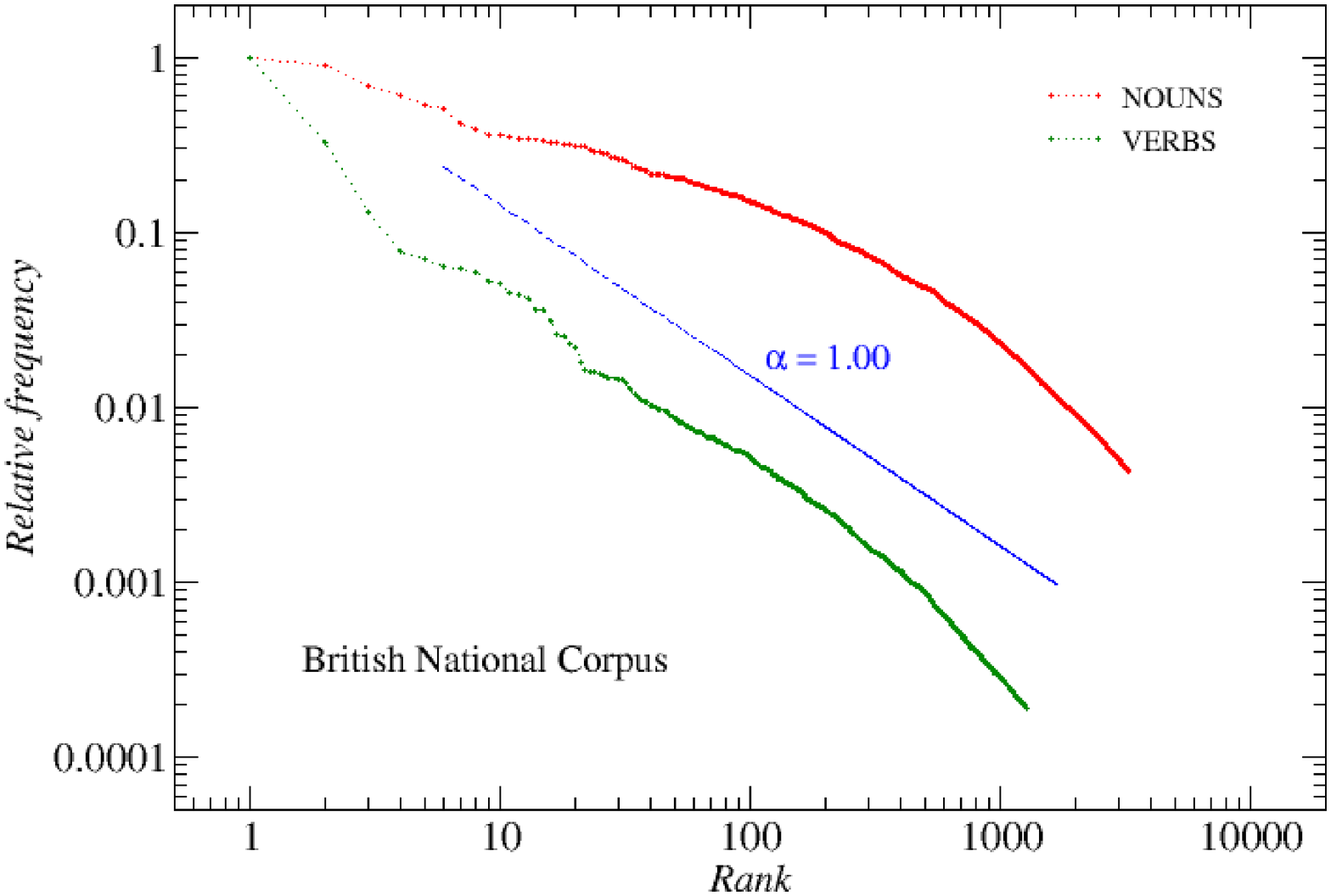}
\caption{Rank-frequency distributions of the most frequent lemmatized
nouns (red) and verbs (green) taken from the British National Corpus. A 
slope with exponent $\alpha = 1.00$ is shown as a benchmark.}
\end{figure}

Any text, as a piece of human syntactic communication, is not a series of
grammatically equivalent words, but rather a convoluted mixture of words
belonging to different parts of speech (classes). Tagging all the words
with their proper parts of speech allows us to compare statistical
properties of words within each class separately. Figure 2 shows the
rank-frequency distributions of the most frequent nouns, verbs,
adjectives, and adverbs in the British National Corpus~\cite{bnc} which is
a representative sample of the contemporary written and spoken English.
The data comes from~\cite{leech}. As it can be seen, despite the fact that
the whole corpus exhibits a Zipf-type scaling for ranks up to several
thousands~\cite{ferrer01}, the corresponding distributions may not
necessarily show any scaling if only words representing a single part of
speech are considered. Verbs are the only class of words that develop some
trace of scaling behaviour with the scaling index $\alpha \approx 1$.
Looking from this perspective at the global distribution of all the words
belonging to all the classes together it is extremely interesting to 
reiterate that the entire rank-frequency distribution is Zipf-like.

This result can reflect a highly convoluted syntactic organization of
human language, which may be considered as a complex system. From this
angle, the linguistic complexity primarily manifests itself through the
logic of mutual dressing among words belonging to different parts of
speech that the entire proportions emerge scale-free even though in
majority of these parts separately the proportions do not respect such a
kind of organization. Another interesting issue open for speculation is
whether the above results actually reflect the fact that the Zipf's
principle of least action is more applicable to verbs - a part of speech
related to action - than to nouns that are linked to objects. Worth
considering is also a possibility that this reflects mapping of the
well-established physical principle of least action onto the frequency
distribution of verbs in a text.

\begin{figure}[t]
\hspace{2.0cm}
\epsfxsize 8cm
\epsffile{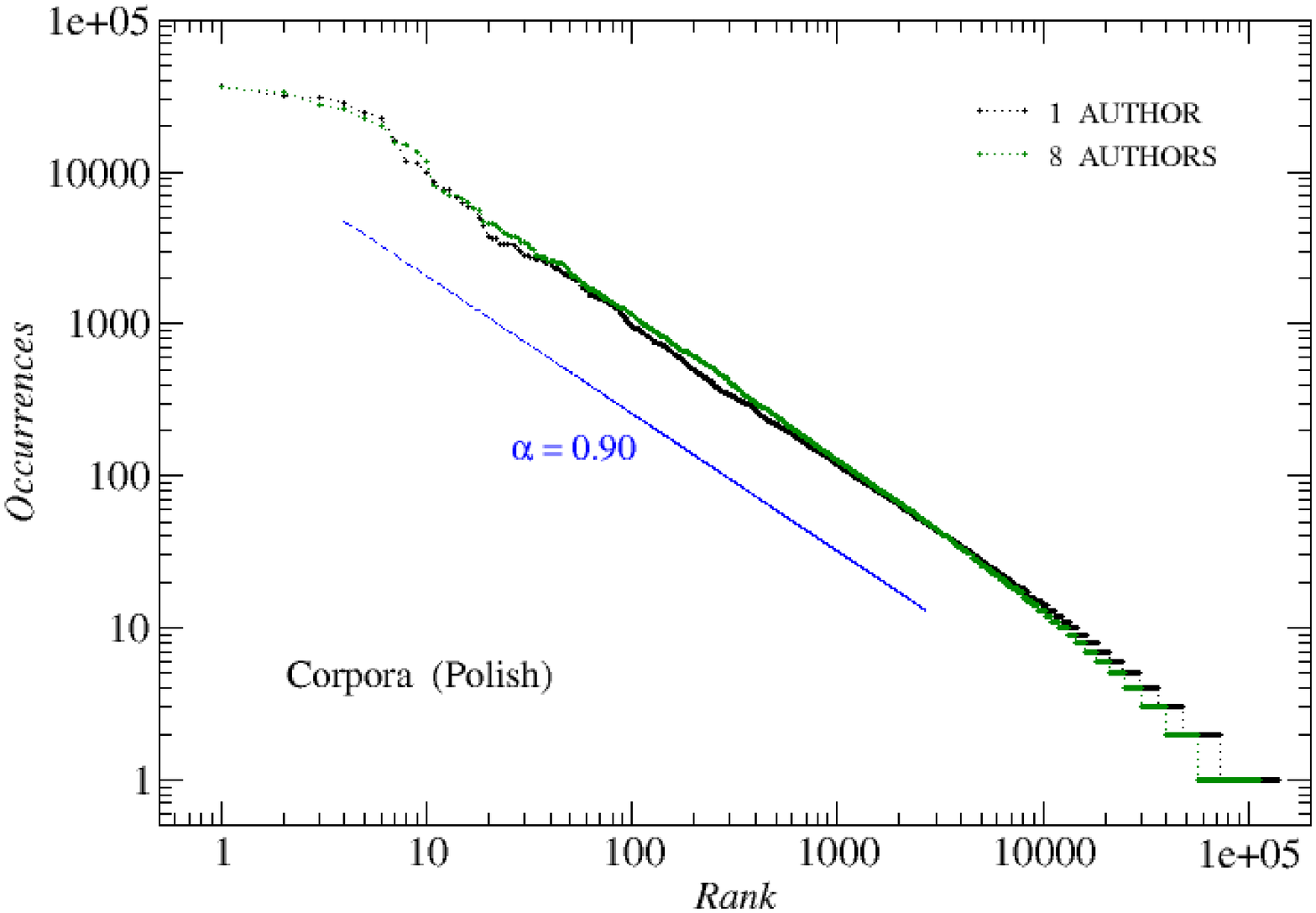}
\caption{Rank-frequency distributions of words for two Polish corpora 
consisting of works of the same author (black) and of different authors 
(green). A power-law with best-fitted exponent $\alpha \simeq 0.90$ is 
denoted by blue dashed line.}
\end{figure}

As it was already mentioned above, passing from a single English text to a
larger literary corpora is associated with reaching the limits of the
applicability of the Zipf law in its classical form with the scaling index
$\alpha = 1$. Typically, after a short transient, for ranks larger than a
few thousands another scaling regime is observed with $\alpha >
1$~\cite{montemurro01,ferrer01}. Presence of two distinct scaling
exponents in the rank-frequency distribution of English words can be
explained by the existence of two sets of words: the first one comprises
common words which are frequently used by all the authors (thus forming
the language core), and the second one comprises the remaining words among
which are technical words, words typical for a specific author or words
which are otherwise rarely used. However, we propose another complimentary
explanation of the breaking of the Zipf law for higher word ranks. Based
on books of a few different authors we observe that the Zipf law is better
realized for single texts than it is for corpora, even if we consider a
corpus to be a collection of works of the same author. Figure 3 shows the
rank-frequency distributions of words for two small corpora of Polish
texts: the first one (black symbols) comprises 26 novels and stories by
Polish fantasy writer Andrzej Sapkowski, while the second one (green
symbols) is formed of 41 novels and stories written by 8 different
authors. The texts in the second corpus were selected in such a way that
the total length of each corpus is comparable (1.3 million words). It can
easily be noted that for ranks larger that 6,000-8,000 the distribution
for the second corpus shows a slightly faster decay than the distribution
for the first corpus. In turn, the distribution for the first corpus seems
to deviate from the unique scaling behaviour more than any single member
text of this corpus (not shown here, but compare this with the result for
a single novel in Figure 1(c)). This conclusion, however, must be treated
with care since single texts have much smaller vocabulary than larger
corpora.

The above evidence suggests that the long-range correlations originating
from a given book's continuous narration can be a strong source of scaling
behaviour. These correlations are distorted if we form a corpus consisting
of different works, in the same manner as the correlations which are
allowed to exist in each particular realization of a system are suppressed
if one forms a statistical ensemble from a number of different
realizations of this system. This possibility opens space for contesting
the traditional model of analysis in quantitative linguistics, according
to which the corpora, due to their larger size, are more useful subject of
analysis than single works. In our opinion this leads to losing a
significant amount of information. The above outcome is another argument
in favour of the concept that the words extracted from their context are
rather different objects even from a purely statistical point of view than
the same words embedded in a contextual environment.

\begin{figure}[t]
\hspace{2.0cm}  
\epsfxsize 8cm  
\epsffile{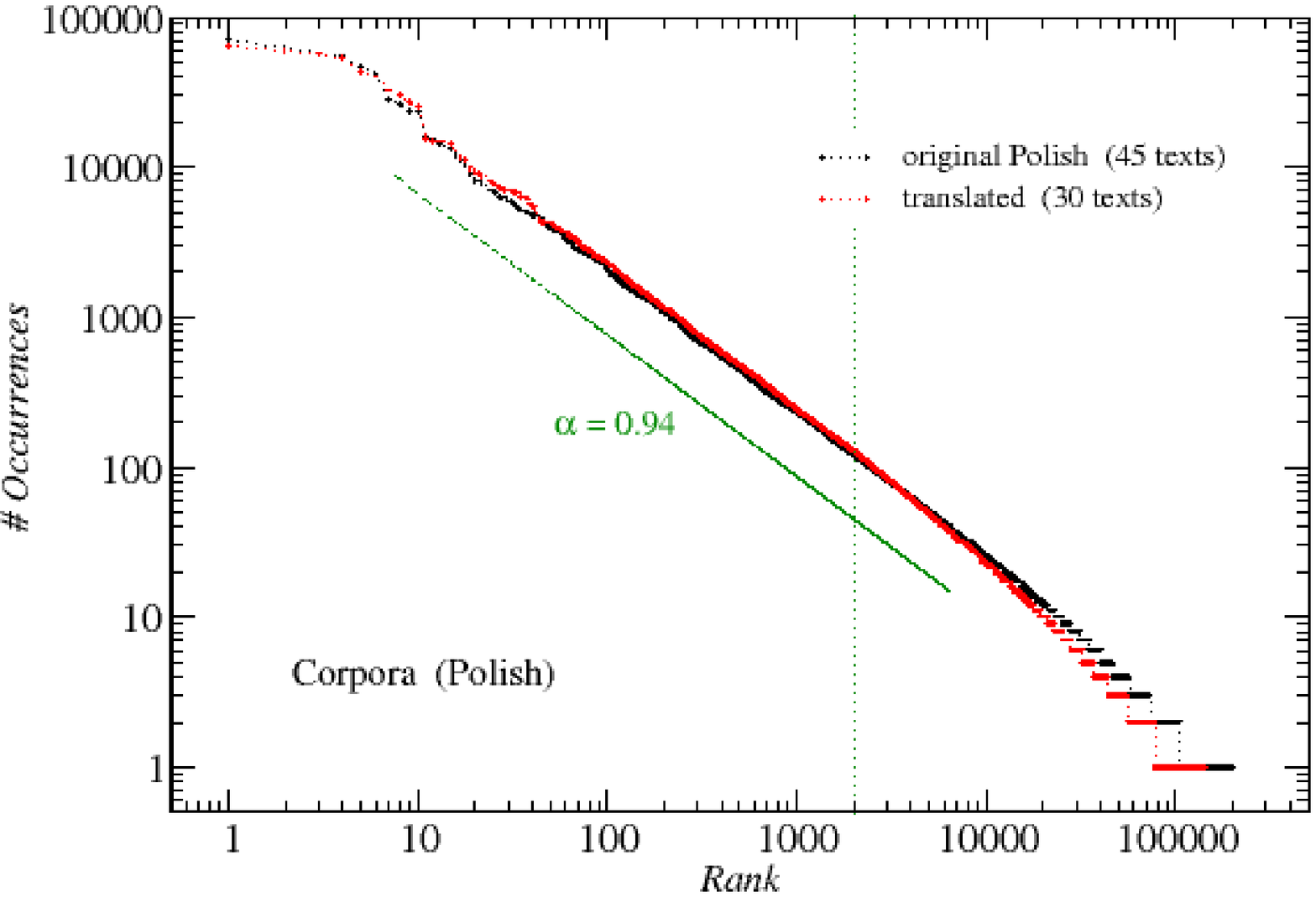}
\caption{Rank-frequency distributions of words for two Polish
corpora consisting of 45 original texts written in Polish (black) and of
30 foreign texts translated into Polish (red). A power law with exponent
$\alpha \simeq 0.94$ well-fitting both distributions within the ranks  
10-2000 is denoted by green dahed line.}
\end{figure}

Finally, let us look once again at Figure 1(a) and Figure 1(b), where the 
Zipf plots for an English text and its Polish translation are presented. 
Both distributions show the undistorted power-law slope for the whole 
range of ranks, which means that the scale-free character of ``Ulysses'' 
was preserved by the translator. Actually, if one takes into account the 
peculiar character of this novel, especially the unequally rich 
vocabulary, this result has to be considered remarkable. Motivated by this 
observation, we more systematically look at the rank-frequency 
distributions of words from texts which were translated into Polish from 
other languages. We find that although such texts show scale-free 
behaviour for the smallest ranks, for larger ones a breakdown of scaling 
occurs and we see a deflection towards smaller frequencies. In order to 
compare scaling properties of the translated and the native Polish texts, 
we constructed the following two small corpora: the first one consisting 
of 45 texts written originally in Polish, and the second one consisting of 
30 translated texts. Both corpora have similar size of 2.3 million words. 
The results are presented in Figure 4. For the ranks $< 2000$ both corpora 
develop roughly the same distributions with $\alpha \simeq 0.94$. However, 
for higher ranks the translated corpus is represented by less frequent 
words than the native corpus. The observed difference is sufficiently 
significant to consider it as an actual property of both considered groups 
of texts. This result does not seem to be unexpected. Generally, the 
higher is $\alpha$ for a sample, the poorer is the vocabulary of a 
corresponding author. It seems natural to expect that vocabulary of a 
writer is richer than a vocabulary of a translator. The first one works 
under no lexical constraints, while the second one has to concentrate 
principally on preserving the sense and the style of the original, what 
can lead to impoverishment of the lexicon. Moreover, differences in 
grammar also may play a role here.

The famous statement of P.W.~Anderson~\cite{anderson}, "More is different",
seems particularly adequate in relation to the human language, indeed.


\begin{thebibliography}{99}

\bibitem{nowak00} M.A.~Nowak, J.B.~Plotkin, V.A.A.~Jansen, The evolution
of syntactic communication, Nature 404, 495-498 (2000)

\bibitem{zipf49} G.K.~Zipf, {\it Human behavior and the principle of least
effort}, Addison-Wesley (Cambridge, 1949)

\bibitem{bnc} The British National Corpus website:
http://www.natcorp.ox.ac.uk/

\bibitem{leech} G.~Leech, P.~Rayson, A.~Wilson, Word Frequencies in
Written and Spoken English: based on the British National Corpus, Longman
(London 2001)

\bibitem{ferrer03} R.~Ferrer Cancho, R.V.~Sol\'e, Least effort and the  
origins of scaling in human language. Proc.~Natl.~Acad.~Sci.~USA 100,
788-791 (2003)

\bibitem{miller57} G.A.~Miller, Some effects of intermittent silence,
Amer.~J.~Psychol. 70, 311-314 (1957)

\bibitem{slomczynski} J.~Joyce, Ulisses, translated by
M.~S{\l}omczy\'nski, Wydawnictwo Pomorze (Bydgoszcz 1992)

\bibitem{montemurro01} M.A.~Montemurro, Beyond the Zipf-Mandelbrot law in
quantitative linguistics, Physica A 300, 567-578 (2001)

\bibitem{ferrer01} R.~Ferrer Cancho, R.V.~Sol\'e, Two regimes in the
frequency of words and the origins of complex lexicons: Zipf's law
revisited, J.~Quant.~Linguistics 8, 165-173 (2001)

\bibitem{anderson} P.W.~Anderson, More is different, Science 177, 393-396
(1972)

\end{thebibliography}
\end{document}